\documentclass[conference]{IEEEtran}
\IEEEoverridecommandlockouts
\usepackage{amsmath,amssymb,amsfonts}
\usepackage{algorithmic}
\usepackage{graphicx}
\usepackage{textcomp}
\usepackage{xcolor}
\def\BibTeX{{\rm B\kern-.05em{\sc i\kern-.025em b}\kern-.08em
    T\kern-.1667em\lower.7ex\hbox{E}\kern-.125emX}}

\usepackage{filecontents}
\usepackage[noadjust]{cite}

\begin{document}

\title{Adaptive Bayesian Linear Regression for Automated Machine Learning
}

\author{\IEEEauthorblockN{ Weilin Zhou}
\IEEEauthorblockA{\textit{I3S laboratory} \\
	\textit{Universite Cote d'Azur}\\
	Nice, France \\
	zhouwl123@gmail.com}
\and
\IEEEauthorblockN{Frederic Precioso}
\IEEEauthorblockA{\textit{I3S laboratory} \\
	\textit{Universite Cote d'Azur}\\
	Nice, France \\
	frederic.precioso@unice.fr}
}

\maketitle

\begin{abstract}
To solve a machine learning problem, one typically needs to perform data preprocessing, modeling, and hyperparameter tuning, which is known as model selection and hyperparameter optimization.
The goal of automated machine learning (AutoML) is to design methods that can automatically perform model selection and hyperparameter optimization without human interventions for a given dataset. 
In this paper, we propose a meta-learning method that can search for a high-performance machine learning pipeline from the predefined set of candidate pipelines for supervised classification datasets in an efficient way by leveraging meta-data collected from previous experiments. 
More specifically, our method combines an adaptive Bayesian regression model with a neural network basis function and the acquisition function from Bayesian optimization. 
The adaptive Bayesian regression model is able to capture knowledge from previous meta-data and thus make predictions of the performances of machine learning pipelines on a new dataset. 
The acquisition function is then used to guide the search of possible pipelines based on the predictions.
The experiments demonstrate that our approach can quickly identify high-performance pipelines for a range of test datasets and outperforms the baseline methods.
\end{abstract}


\section{Introduction}
Modern data analysis for datasets from real world needs complex data processing and modeling. Designing a machine learning model with optimal performance for a given dataset involves in model selection (e.g. which model is best for the task, SVM, Gradient Boosting, or Neural Networks?) and hyperparameter optimization (e.g. Choosing the number of trees in Random Forest, and the number of hidden layers for a neural network model), which was traditionally performed by experts with the trial and error method or simple automatic search methods, e.g. grid search, random search \cite{bergstra2012random}. Recent advances aiming at boosting machine learning result in a research field of automated machine learning (AutoML) \cite{guyon2016brief}\cite{feurer2015efficient}, whose goal is to make machine learning more accessible and user-friendly especially for non-experts and every business \cite{li2018cloud} by automatic model selection and hyperparameter optimization without human intervention. Recent work has been in a wide range from developing advanced optimization methods for hyperparameter optimization, e.g. Bayesian optimization \cite{feurer2015efficient}\cite{li2017hyperband} to neural architecture search through reinforcement learning \cite{45826}\cite{baker2017designing}\cite{Zoph2018LearningTA} and evolutionary algorithms \cite{pmlr-v70-real17a}\cite{xie2017genetic}.

A multi-step process is usually necessary to design a machine learning pipeline with optimal performance for a given dataset, including feature selection, feature construction, model selection, and hyperparameter tuning. 
A typical machine learning pipeline consists of three components, i.e. a feature preprocessor, a machine learning model, and their hyperparameters \cite{feurer2015efficient}. The automatic design of machine learning pipelines has been regarded as a combined model selection and hyperparameter optimization problem \cite{feurer2015efficient}\cite{kotthoff2017auto} and can be formalized as an optimization problem,
\begin{equation}
{\arg \min}_{m,p,\theta_{m},\theta_{p} }\mathcal{L}(m(p(\mathbf{X};\theta_p),\textbf{y};\theta_m))
\end{equation}
where $p\in \mathcal{P}$ and $m\in \mathcal{M}$ denotes a feature preprocessor with hyperparameter $\theta_p$ and a machine learning model with hyperparameter $\theta_m$ respectively, and $\mathcal{P}$ and $\mathcal{M}$ denotes the set of candidate preprocessing methods and machine learning models. $\mathcal{L}$ represents the loss function used to evaluate performance of the pipeline given the test dataset $\mathcal{D}=(\mathbf{X},\mathbf{y})$. 
The hyperparameter space to be optimized, denoted by $\Theta=(m,p,\theta_m,\theta_p)$, is a combined and hierarchical space with discrete, categorical, and continuous dimensions.
In work \cite{feurer2015efficient}, the AutoML problem illustrated by equation (1) was regarded as a hierarchical hyperparameter optimization problem and tackled with Sequential Model-Based Optimization (SMBO) \cite{hutter2011sequential}, also known as Bayesian Optimization (BO), which has been widely adapted for AutoML problem \cite{snoek2012practical}\cite{Liu2018ProgressiveNA}\cite{wang2016bayesian}. 
In the BO setting, the performance of machine learning pipelines is modeled as a function of hyperparameter settings, and a surrogate model, e.g. Gaussian Process or Random Forest, is used to model the function.
The surrogate model is capable of modeling the posterior distribution over the hyperparameter space given a set of observations and then used to select the most promising hyperparameter configuration by trading off exploration of new regions of the space and exploitation in known good regions. 
Next, BO updates the surrogate model with the new observation and then iterates until it runs out of budget (e.g. time).
Bayesian optimization techniques can help to identify better hyperparameters than human experts and have achieved state-of-the-art performance in some applications \cite{feurer2015efficient}\cite{snoek2012practical}.
However, the main drawback of Bayesian optimization techniques is that it suffers from the hierarchical and high-dimensional hyperparameter space. This limitation is due to the necessity of sampling large amounts of hyperparameter settings to model the posterior distribution over the hyperparameter space \cite{wang2016bayesian}, which makes the optimization inefficient.  

In this paper, our goal is to search for the optimal machine learning pipeline for a given dataset from a predefined set of candidate pipelines rather than from the combined, hierarchical, and high-dimensional hyperparameter space.
The predefined set of pipelines is obtained by random sampling from the continuous hyperparameter space. The effectiveness of this random sampling is due to the low effective dimensionality of the hyperparameter space, which has been illustrated in previous work \cite{bergstra2012random}.
We propose to combine an adaptive Bayesian linear regression (ABLR) with a neural network basis and the acquisition function used in BO to deal with the AutoML problem by leveraging previous experiments.
Traditionally, Gaussian Process regression is used as the surrogate model in BO for its capability of capturing the well-calibrated uncertainty. However, it takes cubic time with respect to the number of observations, which makes GP regression unscalable in our multi-task setting containing large numbers of observations.
The complexity of ABLR is linear and the experiments demonstrated that the simple Bayesian linear regression but with a neural network basis performs well in our task.
The input of the ABLR model is a combination of meta-features \cite{feurer2015efficient}, i.e. characteristics of the datasets such as the number of instances, features, and classes, and pipeline embeddings learned while training the basis function. 
To obtain the training data for our method, a large number of machine learning pipelines are configured in advance and constitute the predefined set of pipelines. For example, a configured pipeline can be composed of a PCA model with 6 principal components and a random forest with 500 trees.
The performances of configured pipelines across different datasets were obtained from extensive experiments in which the configured pipelines were tested on hundreds of different datasets.


We test the proposed method and compare the experimental results with two strong baseline methods and Auto-sklearn \cite{feurer2015efficient}, which is a well-known AutoML system and was implemented based on Bayesian optimization. The experiment demonstrates that the proposed method outperforms all the others on our test cases.

\section{Background and Related Work}

\subsection{Bayesian Optimization for AutoML}

Bayesian optimization is a black-box optimization framework that is especially suitable for functions that need expensive evaluations. The goal of BO is to find an optimal solution $x^*$ such that $x^*={\arg \min}_{x\in \mathcal{X}}f(x)$. It is usually expensive to evaluate the function $f(x)$ given an input $x$. Thus, it is important to utilize the surrogate model, such as Gaussian Process, Random Forest, to learn the distribution $p(f|\mathcal{D})$ by some observations. The first step is to use observations $\mathcal{D}=\{(x_1,y_1),(x_2,y_2),\dots,(x_I,y_I)\}$ to initialize the surrogate model. Then, the optimization process performs the following iterations: (1) The surrogate model is trained using current observations $\mathcal{D}_t=\{(x_1,y_1),(x_2,y_2),\dots,(x_t,y_t)\}$ to capture the posterior distribution $p(f|\mathcal{D})$; (2) To select the next observation $(x_{t+1},y_{t+1})$, acquisition function is used by trading off the exploration and exploitation. The new observe data is $\mathcal{D}_{t+1}:=\mathcal{D}_t \cup \{(x_{t+1},y_{t+1})\}$.

Recently, Bayesian optimization has been widely used for AutoML, including works in hyperparamter optimization \cite{snoek2012practical}, pipeline search \cite{kotthoff2017auto}\cite{feurer2015efficient}, neural architecture search \cite{Liu2018ProgressiveNA}. The very related work to this paper is \cite{feurer2015efficient}, in which a machine learning pipeline is composed of one data preprcessor, one feature preprocessor, one model, and their hyperparameters. The combined model selection and hyperparameter optimization problem was formalized as a hierarchical hyperparameter optimization with 110-dimensional search space, which was tackled by Bayesian optimization. In the practical problem of AutoML, the input space $x=(m,p,\theta_m,\theta_p)$ is high-dimensional and hierarchical since many candidate models are associated and there some conditional hyperparaters exit in the input space. Besides, to evaluate the preformance of a pipeline with specific hyperparameters, it has to be tested on the held-out dataset which is timely and computationally expensive during the iterations of BO. 
In addition, evolutionary algorithms have also been used for pipeline optimization \cite{olson2016evaluation}. Genetic programming is used to optimize the pipelines for a given dataset by encoding the samples of pipeline with genes. However, a big population composed of large amounts genes is required for the evolution based optimization since this method uses the similar hyperparameter space as used in BO methods. 
Recent work \cite{NIPS2018_7595} leveraged meta-data from previous experiments to learn a latent representation of pipelines using the probabilistic matrix factorization and the acquisition function is then used to guide the search for an optimal pipeline in the latent space for a new dataset. 
Our work also explores the meta-data collected from the previous experiments, but different from work \cite{NIPS2018_7595} the proposed method utilizes the adaptive Bayesian linear regression with a neural network as the basis to learn a distribution of performance conditional on pipeline embeddings and mete-features of the test datasets. 
A multi-task ABLR model has been proposed in previous work \cite{NIPS2018_7917}, while it focuses on hyperparameter transfer learning for a single machine learning model in BO.

\subsection{Bayesian methods for neural networks}
The goal of the optimization in AutoML is to search for the pipeline that gives the best performance for a dataset. In the setting of Bayesian optimization for AutoML, the performance is seem as a function of the input hyperparameters, and the surrogate model is used to model the distribution $p(f|\mathcal{D})$. As we mentioned before, the high-dimensional and hierarchical input hyperparameter space leads to an inefficient optimization. In order to release this issue, the performance $f$ is directly regarded as a function of the dataset represented by meta-features and the pipelines represented by the embeddings. Neural networks is a natural choice to model the distribution $p(f|\mathcal{D})$ for its flexibility and capability of constructing the learnable embeddings. The BO benefits from the well-calibrated uncertainty estimates of the posterior distribution given by the surrogate model, such as Gaussian Process. To this end, Bayesian neural network is a good choice for the AutoML task. 

The goal of Bayesian methods for neural networks is to model the posterior distribution over the network weights, which has been explored by researchers for a long history, including early work on Hamiltonian Monte Carlo \cite{neal2012bayesian}, recent work on expectation propagation \cite{hernandez2015probabilistic}, and approximate inference \cite{gal2016dropout}\cite{NIPS2015_5666}, as well as stochastic gradient Hamiltonian Monte Carlo \cite{chen2014stochastic}. However, in practice these methods require expensive Markov Chain Monte Carlo simulation or approximate inference, which makes it difficult to deploy these methods in our solution to the AutoML problem.
Therefore, in this work we adopted the Bayesian linear regression model with a neural network basis to model the distribution of the performance given the pipelines and meta-features.

\section{ABLR for AutoML}
The meta-data that will be used for training the adaptive Bayesian linear regression model is denoted by $Y \in \Re^{N\times D}$, where $N$ and $D$ represents the number of pipelines and datasets considered in the previous experiments, and the element of $Y$ denoted by $y_{i,j}$ is the balanced accuracy of the pipeline $i$ tested on the held-out data of the dataset $j$. We will use a feed-forward neural network model with an embedding layer to learn an embedding for each pipeline. Thus, each pipeline is associated with an embedding $\psi_i$. Besides, each dataset used in the experiments is associated with a meta-feature vector denoted by $f_j$ containing features such as the number of instances, the number of features in the dataset. 

\subsection{Adaptive bayesian linear regression with neural networks}\label{AA}
 In the paper, we will use the adaptive Bayesian linear regression model \cite{anzai2012pattern} with a feed-forward neural network model as the basis function to capture the uncertainty of the performance with respect to pipelines and the meta-features of a dataset. Therefore, the adaptive Bayesian linera model allows for searching for an optimal pipeline for give dataset by explicitly trading off the exploration and exploitation.
The key work for the ANLR model is to design a basis function that can transfer the input features to a proper representation for the Bayesian linear regression model. Inspired by the work \cite{snoek2015scalable}, we use a neural network as the basis function.

The basis function is obtained by training a neural network mode with an embedding layer, which can be represented as
\begin{equation}
	\varphi(f_j,i;\theta)=h([f_j;\psi_i]^T,\theta)
\end{equation}
where $f_j$ is the meta-features of dataset $j$, and i is an indicator of a pipeline. $\psi_i$ is the i-th row of the embedding matrix $\psi \in \Re^{N\times L}$ ($L$ is the dimension of the pipeline embedding set manually in the following experiment). The function $h$ is implemented using a feed-forward neural network with parameters $\theta$. Thus, the network parameter is $\Theta=[\theta, \psi] $, which is learned by training on input and target data $\mathcal{D}=\{(x_{i,j},y_{i,j})|1\leqslant i\leqslant N, 1\leqslant j\leqslant D, y_{i,j}\in Y\}$, where $x_{i,j}=(f_j,i)$, with squared loss function and stochastic gradient descent. The outputs from the last hidden layer of the neural network is taken as the basis function, denoted by $\phi=[\phi_1,\phi_2,\cdots,\phi_M]^T$, assuming that the size of the last hidden layer is $M$. 
Thus, the design matrix can be defined by $\Phi$, in which $\Phi_{(i,j),m}=\phi_{m}(x_{i,j})$.
The basis function is parameterized via weights $\Theta$ of the neural network model except the weights of the last linear layer.

From the perspective of Bayesian linear regression, we have
\begin{equation}
	y_{i,j}|x_{i,j},\mathbf{w},\alpha, \Theta \sim \mathcal{N}(\mathbf{w}^T\phi(x_{i,j}),\alpha^{-1})
\end{equation}
A prior Gaussian distribution with zero-mean and precision $\beta$ is imposed on parameters $\mathbf{w}$. Thus, the posterior distribution over parameters $\mathbf{w}$ is obtained. The predictive distribution for a new input $x_{n,d}^*$ for pipeline $n$ and dataset $d$ can be inferred analytically \cite{anzai2012pattern} and the mean $u(x_{n,d}^*)$ and the variance $\sigma^2(x_{n,d}^*)$ is 
\begin{align}
	\mu(x_{n,d}^*;\mathcal{D},\alpha,\beta,\Theta)&=\mathbf{m}^T \phi(x_{n,d}^*)\\
	\sigma^2(x_{n,d}^*;\mathcal{D},\alpha,\beta,\Theta)&=\phi(x_{n,d}^*)^T \mathbf{K}^{-1} \phi(x_{n,d}^*)
\end{align}
where
\begin{align}
	\mathbf{m}&=\beta \mathbf{K}^{-1}\Phi^T\mathbf{y}\\
	\mathbf{K}&=\beta\Phi^T\Phi+\mathbf{I}\alpha
\end{align}
Thus, the log-marginal likelihood is given by
\begin{align*}
\log p(\mathbf{y},\alpha,\beta)&=\frac{M}{2} \log \alpha + \frac{Q}{2} \log \beta -  \frac{Q}{2} \log (2\pi) \\
&-\frac{\beta}{2}\lVert\mathbf{y}-\Phi\mathbf{m}\rVert^2 - \frac{\alpha}{2} \mathbf{m}^T \mathbf{m}-\frac{1}{2}\log \lvert\mathbf{K}\rvert
\end{align*}
where $Q$ denotes the number of instances in the data $\mathcal{D}$.
The main computational load is to invert the matrix $K$. The inversion takes linear time with respect to the number of observations although it scales cubicly in the output dimensionality of the basis function. 
The standard empirical Bayes approach to estimate the parameter $\alpha,\beta$ and $\Theta$ is to maximize the log-marginal likelihood using gradient-based optimizer. However, each step of update of stochastic gradient descent involves in inverting the matrix $K$, which makes the optimization significantly slow. Hence, we adopted a point estimate method to estimate the parameter $\Theta$. As suggested by the previous work \cite{snoek2015scalable}, the optimization consists of two steps. The first one is to optimize $\Theta$ by minimizing a squared loss function after adding a linear output layer to the basis function. Next, $\alpha,\beta$ are estimated by maximizing the log-likelihood function.
\subsection{Acquisition functions}
We have designed an adaptive Bayesian linear regression model by leveraging the previous experiments. For a new dataset, the regression model can provide predictive means and variances of performances for pipelines considered in the experiments. Therefore, the regression model will be used to improve the search efficiency for the optimal pipeline. In the search iterations, a pipeline to be evaluated is chosen based on the acquisition function. A simple method is to select the pipeline with the maximum predictive mean, i.e. $argmax_n (u(x_{n,d}))$. However, this approach do not take the uncertainty into consideration. A commonly used acquisition function in Bayesian optimization is Expectation Improvement \cite{brochu2010tutorial}, which is given by
\begin{equation*}
EI_{n,d}=\mathop{\mathbb{E}} [(y_{n,d}-y_{best}) \mathop{\mathbb{I}} (y_{n,d}>y_{best})]
\end{equation*}
where $y_{best}$ is the best performance observed. Since $y_{n,d}$ is Gaussian distributed, the expected improvement criterion can be obtained analytically:
\begin{equation*}
EI_{n,d}=\sigma(x_{n,d})[\gamma(x_{n,d}) \Phi( \gamma(x_{n,d}) )  +\mathcal{N}( \gamma(x_{n,d});0,1 )]
\end{equation*}
\begin{equation*}
	\gamma(x_{n,d})=\frac{\mu(x_{n,d})-y_{best}-\xi}{\sigma(x_{n,d})}
\end{equation*}
where $\Phi(\cdot)$ denotes the cumulative distribution function of a standard normal, and $\mathcal{N}(\cdot;0,1)$ is the density function of a standard normal. $\xi$ is a parameter used to control the trade-off of exploration and exploitation. Thus, the next pipeline selected to evaluate is given by $argmax_n (EI_{n,d})$.

\section{Experiments}
\subsection{Training data}
We collected 100 datasets for classification from the UCI repository \cite{Dua-2017}. The number of instances in these datasets is in the range from five hundred to fifty thousand. Nearly 1000 pipelines are sampled from the candidate set of preprocessing methods $\mathcal{P}=\{p_1,p_2,\cdots,p_m\}$ with the corresponding hyperparameters $\Theta_p=\{\theta_1,\theta_2,\cdots,
\theta_m\}$ and models $\mathcal{M}=\{m_1,m_2,\cdots,m_n\}$ with the corresponding hyperparameters $\Theta_m=\{\theta_1,\theta_2,\cdots,\theta_n\}$.
All of these pipelines are evaluated on the 100 datasets respectively and the classification accuracy is used as the performance metric. During the evaluations, each dataset was split in 80\% training data, 10\% validation data, and 10\% test data.
We use the implementations of preprocessing methods and models in Weka \cite{hall2009weka} to run the evaluations. Note that some tests may fail due to time and memory limit.
The experiments generated around 54,000 results, which will be used as the meta-data.
Besides, the training data contains the meta-features of the datasets. As performed in the previous work \cite{feurer2015initializing}, the meta-feature includes 47 attributes, such as \textit{number\_of\_instances}, \textit{number\_of\_attributes}, \textit{class\_entropy}, \textit{coefficient\_of\_variation}. All of the attributes of the meta-feature used in the training data will be listed in the appendix. The training data comprises the performance results of the pipelines on the 100 datasets and the meta-features of the datasets. Among the training data, the data from 70 datasets is used for training of the adaptive Bayesian regression model, while the left is used for test.

\subsection{Model details}
The proposed adaptive Bayesian regression model aims at learning a performance distribution conditional on datasets and pipelines. In the model, the datasets and pipelines are characterized by meta-features and pipeline embeddings respectively.
Training such model consists of two phrases. 
The first phrase is to train a feed-forward neural network model, which will be used for the basis function of the Bayesian regression model. The neural network model is trained using the meta-data $\mathcal{D}=\{(x_{i,j},y_{i,j})|x_{i,j}=(f_j,i)\}$, where $f_j$ denotes the meta-feature of dataset $j$ and $i$ is the indicator of pipeline $i$. 
The input feature of the neural network model is a combination feature of meta-features and pipelines. In our model, the meta-feature is a 47-dimensional vector and the pipeline is a learnable embedding with $L$ dimensions. 
The target of the neural network model is the corresponding performance data, i.e. classification accuracy $y_{i,j}$.
However, training the neural network model involves in tuning hyperparameters including the dimension $L$ of the pipeline embedding, the number of hidden layers, the number of neurons in each hidden layer, and the activation function. One can think of tuning these hyperparameters as a hyperparameter optimization problem. We obtained the optimized hyperparameters using the hyperparameter optimization library HPOlib \cite{EggFeuBerSnoHooHutLey13}, which was implemented based on Bayesian optimization. The dimension $L$ of the pipeline embedding is set to 20. The neural network contains five hidden layers with hidden size $(500,200,100,50,50)$. 
Since the work \cite{snoek2015scalable} have shown the empirical results that the commonly used rectified linear function (ReLU) can result in poor estimates of uncertainty, we used the bounded tanh function in our neural network model. The parameters of the neural network model was optimized by backpropagation and stochastic gradient descent with a squared loss function. This procedure can be thought of as a maximum posterior estimate of parameters in the neural network.

\begin{figure}[t]
	\centering
	\includegraphics[width=.5\textwidth]{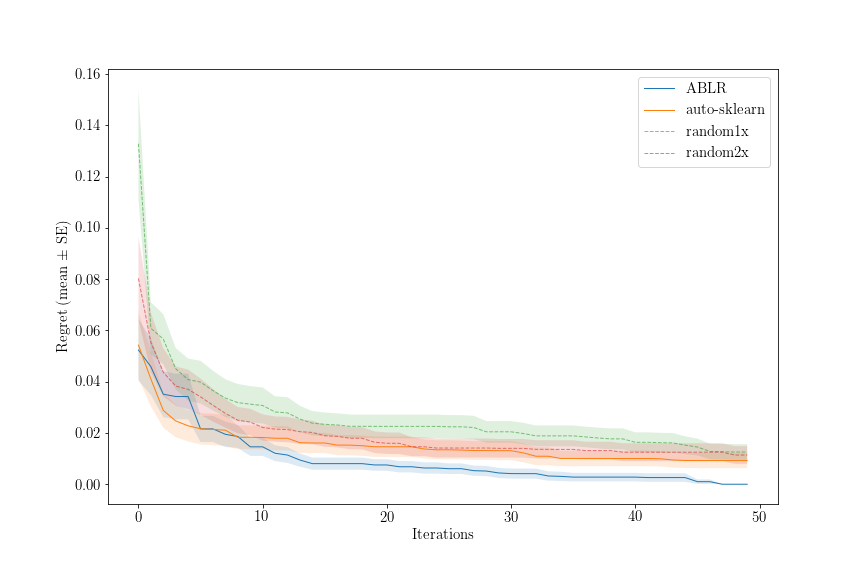}
	\caption{The average regrets of all methods as a function of
		the number of iterations. Each method was evaluated on 30 test datasets. Lower is better. The shaded area represents the standard error for each method.}
	\label{fig1}
\end{figure}

Once the neural network model has been trained, the output layer of the net will be replaced with a Bayesian linear regressor that captures uncertainty in the weights. The second phrase is to estimate the hyperparameters $\alpha$ and $\beta$ by maximizing the log-marginal likelihood function, which was optimized using the L-BFGS algorithm in our experiment. Therefore, the predictive distribution for a new input can be inferred as shown in Equation (4) and (5).

\subsection{Results}\label{SCM}
The training data was generated from 100 datasets, in which the data from 70 datasets was used to train the adaptive Bayesian model and the left was used for evaluating the model performance. In this section, we show the test results.
First of all, we designed four experiments introduced as follows for comparison.
\begin{itemize} 
\item \textbf{Random1x}. The random search is performed from the set of pipelines. Each iteration during the search randomly choose a pipeline to be evaluated on the given dataset. This experiment is regarded as the baseline. 
\item \textbf{Random2x}. The random search is performed with twice budget. It simulates the parallel evaluations of pipelines on the given dataset.
\item \textbf{Auto-sklearn}. We use the open-source implementation of Auto-sklean to run the evaluations and set the time of optimization for each dataset to 30 minutes. All test datasets are evaluated on a holdout dataset. To obtain a fair comparison to other methods, we disabled the automated ensemble construction.
\item \textbf{ABLR}. The proposed adaptive Bayesian regression model with acquisition function (EI) for guiding search. 
\end{itemize}

\begin{figure}[t]
	\centering
	\includegraphics[width=.5\textwidth]{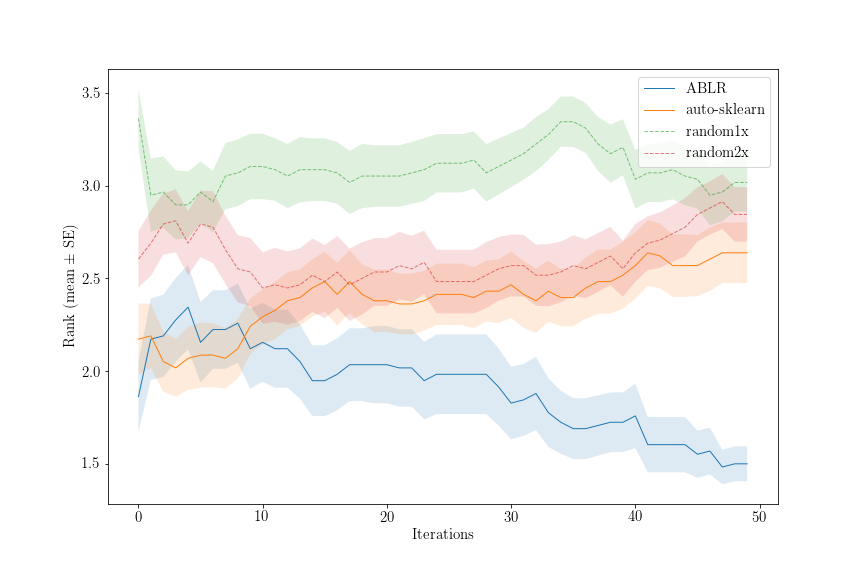}
	\caption{The average ranks of all methods as a function of the number of iterations. Lower is better. Each method was evaluated on 30 test datasets. The shaded area represents the standard error for each method.}
	\label{fig2}
\end{figure}

Using the ABLR model that already have trained on training data, firstly we can obtain a predictive performance of the pipelines on a test dataset. Then, the acquisition function will choose the next pipeline to be evaluated on the test dataset. In the experiments, we set the number of search iterations to 50, and the best performance achieved so far in each iteration is tracked. The regret is defined as the difference between the best classification accuracy achieved in the all pipelines considered in our experiments and the one achieved so far in each iteration. Figure \ref{fig1} shows the average regrets of the four approach evaluated on the 30 test datasets during the search iterations. This figure shows that our method can quickly identify the best pipelines from the set of candidate pipelines. In order to compare the four methods in the experiments more clearly, Figure \ref{fig2} shows the average ranks of the four approach evaluated on all test datasets during the search iterations. This figure shows reasonable results that the Random2x ranks better than Random1x. Besides, The figure also shows the proposed method achieves lowest ranks after few iterations, which demonstrates that our method outperforms the others. 
In Table 1, the best classification accuracies achieved in the 50 iterations of 4 methods on 30 test datasets are listed. We compare the results and note that the ABLR based method achieved the best on 18 datasets out of 30 test datasets.

\begin{table}[]
	\centering
	\caption{Best Test Set Classification Accuracy Achieved in the 50 iterations of 4 Methods on 30 Test Datasets. All of the scores are in percentage. Bold Numbers Indicate the Best Result.}
	\begin{tabular}{ccccc}
		\hline
		Dataset            & Random1x        & Random2x        & Auto-sklearn    & ABLR            \\
		\hline \hline
		Wine-quality-white & \textbf{68.95}  & 68.86           & \textbf{68.95}  & 66.85           \\
		Pima               & \textbf{77.99}  & \textbf{77.99}  & 77.87           & 77.17           \\
		Nursery            & 99.89           & \textbf{99.95}  & 99.91           & 99.42           \\
		Balance-scale      & 91.85           & 90.40            & 92.64           & \textbf{97.1}   \\
		Spect              & \textbf{83.75}           & \textbf{83.75}     & 80.00           & \textbf{83.75}  \\
		Ozone              & 97.24           & \textbf{97.28}  & 97.24           & 94.38           \\
		Mushrooms          & \textbf{100.00} & \textbf{100.00} & \textbf{100.00} & \textbf{100.00} \\
		Musk-1             & 92.43           & 90.31           & \textbf{98.09}           &  93.93 \\
		Spectf             & 83.75           & 82.50            & 82.50            & \textbf{87.50}  \\
		Lymphography       & 85.19           & 85.19           & \textbf{85.90}  & 85.71           \\
		Zoo                & 98.00           & 98.00           & 98.00           & \textbf{99.00} \\
		Seeds              & 96.19           &   \textbf{96.67}         & 95.24           & \textbf{96.67}  \\
		Acute-inflammation & \textbf{100.00} & \textbf{100.00} & \textbf{100.00} & \textbf{100.00} \\
		Glass              & 74.81           & 79.91           &\textbf{85.00}           &  79.91  \\
		Synthetic-control  & 99.00           & 99.17           & \textbf{99.79}     &  99.50\\
		Wine-quality-red   & 68.60            & 68.92           & 69.92           & \textbf{70.04}  \\
		Tic-tac-toe        & 98.43           & 99.58           & 98.96           & \textbf{99.79} \\
		Car                & \textbf{99.6}   & 99.25           & 99.02           & 98.42           \\
		Musk-2             & 99.64           & \textbf{99.74}  & 98.24           & 98.62           \\
		Yeast              & 60.25           & 60.85           & 60.41  &    \textbf{62.26}        \\
		Chess-krvkp        & 99.44           & \textbf{99.59}  & 80.88           & \textbf{99.59}  \\
		Spambase           & \textbf{95.61}  & 94.96           & \textbf{95.61}  & 94.47           \\
		Waveform           & \textbf{87.14}  & \textbf{87.14}  & 86.79  & \textbf{87.14}           \\
		Parkinsons         & 92.37           & 94.84           &  \textbf{96.92}         & 93.32   \\
		Steel-plates       & 78.15           & 77.90            & 78.94           & \textbf{78.99}  \\
		Libras             & 83.61           & 84.72           & 86.55  &  \textbf{87.50}          \\
		Ringnorm           & \textbf{98.68}  & \textbf{98.68}  & 98.32           & \textbf{98.68}  \\
		Adult              & 84.93           & 85.83           & 85.91           & \textbf{86.77}  \\
		Ionosphere         & 93.75           & 94.60            & 94.83           & \textbf{95.43}  \\
		Primary-tumor      & 48.48           & 46.36           & 46.67           & \textbf{49.11}  \\
		\hline
	\end{tabular}
\end{table}

\section*{Conclusion}
We presented a new meta-learning approach to automatically predict high-performance machine learning pipelines for a given dataset. Our method combines the adaptive Bayesian linear regression model with a neural network basis and the acquisition function utilized widely in the Bayesian optimization. We have performed experiments on a set of test datasets, which demonstrated that the proposed method outperforms two strong baseline methods and the well-known AutoML system Auto-sklearn on our test cases. The proposed method performs searching from the discrete set of pipelines and thus avoids direct optimization in the continuous hyperparameter space. 
The experimental results have shown that the scalable ABLR model performs well in our task and searching for an optimal pipeline from the discrete set is a promising direction for pipeline optimization.
In the future, we will work on including more pipelines to increase the possibility of identifying pipelines with higher performance for a given dataset. We will also investigate Bayesian methods for neural networks for the automated machine learning problem.

%
%

\bibliographystyle{IEEEtran}
\bibliography{bibi}

\end{document}